\documentclass[conference]{IEEEtran}
\IEEEoverridecommandlockouts                              % This command is only needed if 
\usepackage{times}
\usepackage[numbers]{natbib}
\usepackage{multicol}

\usepackage{subfig}
\usepackage{url}
\usepackage[dvipsnames]{xcolor}
\usepackage{multirow}
\usepackage{bbold}
\usepackage{balance}

\usepackage{url}

\usepackage{breakurl}

\usepackage{footnote}
\makesavenoteenv{tabular}
\makesavenoteenv{table}

\def\HiLi{\leavevmode\rlap{\hbox to \hsize{\color{yellow!20}\leaders\hrule height .8\baselineskip depth .5ex\hfill}}}
\def\HiLiGreen{\leavevmode\rlap{\hbox to \hsize{\color{green!20}\leaders\hrule height .8\baselineskip depth .5ex\hfill}}}
\def\HiLiBlue{\leavevmode\rlap{\hbox to \hsize{\color{blue!20}\leaders\hrule height .8\baselineskip depth .5ex\hfill}}}
\def\HiLiGray{\leavevmode\rlap{\hbox to \hsize{\color{grey!50}\leaders\hrule height .8\baselineskip depth .5ex\hfill}}}

\usepackage{epsfig}
\usepackage{amssymb}
\usepackage{amsmath}
\usepackage{amsfonts}

\usepackage{stackengine}
\usepackage{algorithmic,comment}
\usepackage[ruled,vlined,commentsnumbered,titlenotnumbered]{algorithm2e}
\usepackage{float}

\usepackage{colortbl}
\definecolor{LightCyan}{rgb}{0.88,1,1}

\usepackage{color}
\newcommand{\SC}[1]{{\color{black}#1}}
\newcommand{\RSS}[1]{{\color{black}#1}}
\newcommand{\tu}[1]{{\underline{\textit{#1}}}}

\newcommand{\name}{\textsc{HarvestNet}\xspace}

\newcommand{\dataset}{\mathcal{D}}
\newcommand{\gretrain}{g_\mathrm{retrain}}
\newcommand{\gevaluate}{g_\mathrm{evaluate}}
\newcommand{\yconf}{\mathtt{conf}}

\newcommand{\fDNNrobot}{f_{\mathrm{DNN}}}

\newcommand{\targetset}{\mathtt{TargetSet}}
\newcommand{\targetimageset}{\mathtt{TargetImageSet}}
\newcommand{\targetclass}{\mathtt{TargetClass}}
\newcommand{\targetclassset}{\mathtt{TargetClassSet}}

\newcommand{\yhat}{\hat{y}}
\newcommand{\pisample}{\pi_{\mathrm{sample}}}

\newcommand{\updatethreshold}{\mathtt{AdaptThresh}}

\newcommand{\boldy}{\boldsymbol{y}}

\newcommand{\Cache}{\mathtt{Cache}}
\newcommand{\Cachesize}{N_{\mathrm{cache}}}
\newcommand{\yground}{y_{\mathrm{GroundTruth}}}

\newcommand{\dtrain}{\mathcal{D}_\mathrm{train}}
\newcommand{\dval}{\mathcal{D}_\mathrm{val}}
\newcommand{\dtest}{\mathcal{D}_\mathrm{test}}
\newcommand{\dtestfinal}{\mathcal{D}_\mathrm{test,~final}}
\newcommand{\roundmax}{N_{\mathrm{round}}}
\newcommand{\rhothresh}{\rho_{\mathrm{thresh}}}
\newcommand{\emb}{\mathtt{emb}}
\newcommand{\yconfthresh}{\yconf_{\mathrm{thresh}}}
\newcommand{\thetasampler}{\theta_{\mathrm{sampler}}}

\newcommand{\argmin}{\mathop{\rm argmin}}

\newcommand{\maskrcnn}{{Faster R-CNN~}}

\newtheorem{problem}{Problem}

\begin{document}
% paper title
%\title{HarvestNet: Mining Valuable Training Data from High-Volume Robot Sensory Streams}

\title{Sampling Training Data for Continual Learning Between Robots and the Cloud}

\author{Sandeep Chinchali$^{1*}$, Evgenya Pergament$^{1*}$, Manabu Nakanoya$^{2}$, 
Eyal Cidon$^{1}$, Edward Zhang$^{3}$, Dinesh Bharadia$^{4}$, \\ 
Marco Pavone$^{1}$, and Sachin Katti$^{1}$     
% <-this % stops a space
\thanks{* Co-Primary Authors}% <-this % stops a space
\thanks{$^{1}$ Depts. of CS, EE, and Aero-Astro, Stanford University, Stanford, USA
{\tt\small \{csandeep, evgenyap, mnakanoya, ecidon, pavone, skatti\}@stanford.edu}}%
\thanks{$^{2}$ NEC Corporation, Tokyo, Japan {\tt\small nakanoya@nec.com}}%
\thanks{$^{3}$ {\tt\small zhang.ed.1998@gmail.com}}%
\thanks{$^{4}$ Dept. of ECE, UC San Diego, La Jolla, USA  {\tt\small dineshb@ucsd.edu}}%
}

\maketitle

\begin{abstract}
    Today's robotic fleets are increasingly measuring high-volume video and LIDAR sensory streams, which
    can be mined for valuable training data, such as rare scenes of road construction sites, to 
    steadily improve robotic perception models. However, re-training perception models
    on growing volumes of rich sensory data in central compute servers (or the ``cloud'') places an enormous time and cost burden
    on network transfer, cloud storage, human annotation, and cloud computing resources. Hence, we introduce \name, an intelligent sampling algorithm that resides on-board a robot and \SC{reduces system bottlenecks} by only storing rare, useful events to steadily improve perception models re-trained in the cloud.
    \name significantly improves the accuracy of machine-learning models on our novel dataset of road construction sites, field testing of self-driving cars, and streaming face recognition, while reducing cloud storage, dataset annotation time, and cloud compute time by between \SC{$65.7-81.3\%$}. 
    Further, it is between \SC{$1.05-2.58\times$} more accurate than baseline algorithms and scalably runs on embedded deep learning hardware.
\end{abstract}

%\IEEEpeerreviewmaketitle
%\maketitle

\section{Introduction}
\label{sec:intro}
\label{sec:intro}

Learning to identify rare events, such as traffic disruptions due to construction or hazardous weather conditions, can significantly
improve the safety of robotic perception and decision-making models. While performing their primary tasks, fleets of future robots, ranging from delivery drones to self-driving cars, can \textit{passively} collect rare training data they happen to observe from their high-volume video and LIDAR sensory streams. 
As such, we envision that robotic fleets will act as \textit{distributed, passive} sensors that collect valuable training data to steadily improve the robotic autonomy stack.

\begin{figure}[t]
	\centering
	\includegraphics[width=1.0\columnwidth]{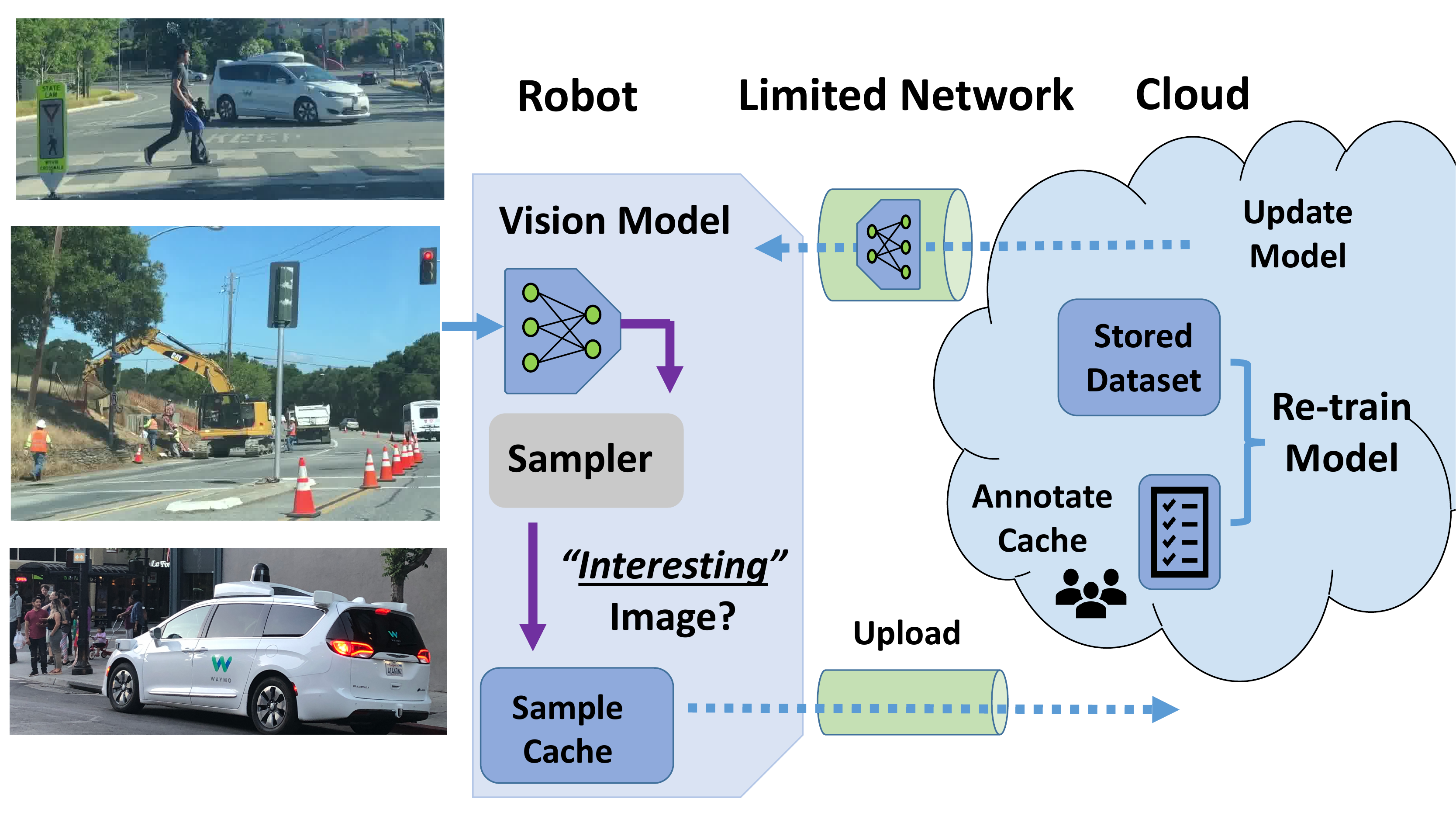}
	\caption{\textbf{The Robot Sensory Sampling Problem.}
    What are the minimal set of ``interesting'' training samples robots can send to the cloud to periodically improve perception model accuracy while minimizing system costs of network transfer, cloud storage, human annotation, and cloud-computing?}
	\label{fig:model_retrain}
\end{figure}

Despite the benefits of continually improving robotic models from field data,
they come with severe systems costs that are largely under-appreciated in today's
robotics literature.
Specifically, these systems bottlenecks (shown in Fig. \ref{fig:model_retrain}) stem from the time or cost of (1) network data transfer to compute servers, (2) limited robot and cloud storage, (3) dataset annotation with human supervision, and (4) cloud computing for training models on massive datasets. Indeed, it is estimated that a \textit{single} self-driving car will produce upwards of 4 Terabytes (TB) of LIDAR and video data per day \cite{intel,tuxera}, which we contrast with our own calculations in Section \ref{sec:motivation}.
%Overcoming systems bottlenecks would allow robots to download improved perception models before the next field test. 

In this paper, we introduce a novel \textit{Robot Sensory Sampling Problem} (Fig. \ref{fig:model_retrain}), which asks how a robot can maximally improve model accuracy via cloud re-training, but with minimal end-to-end systems costs.
To address this problem, we introduce \name, a compute-efficient sampler that scalably runs on even resource-constrained robots and reduces systems bottlenecks by ``harvesting'' only task-relevant training examples for cloud processing. 

\begin{figure}[h]
	\centering
	\includegraphics[width=1.0\columnwidth]{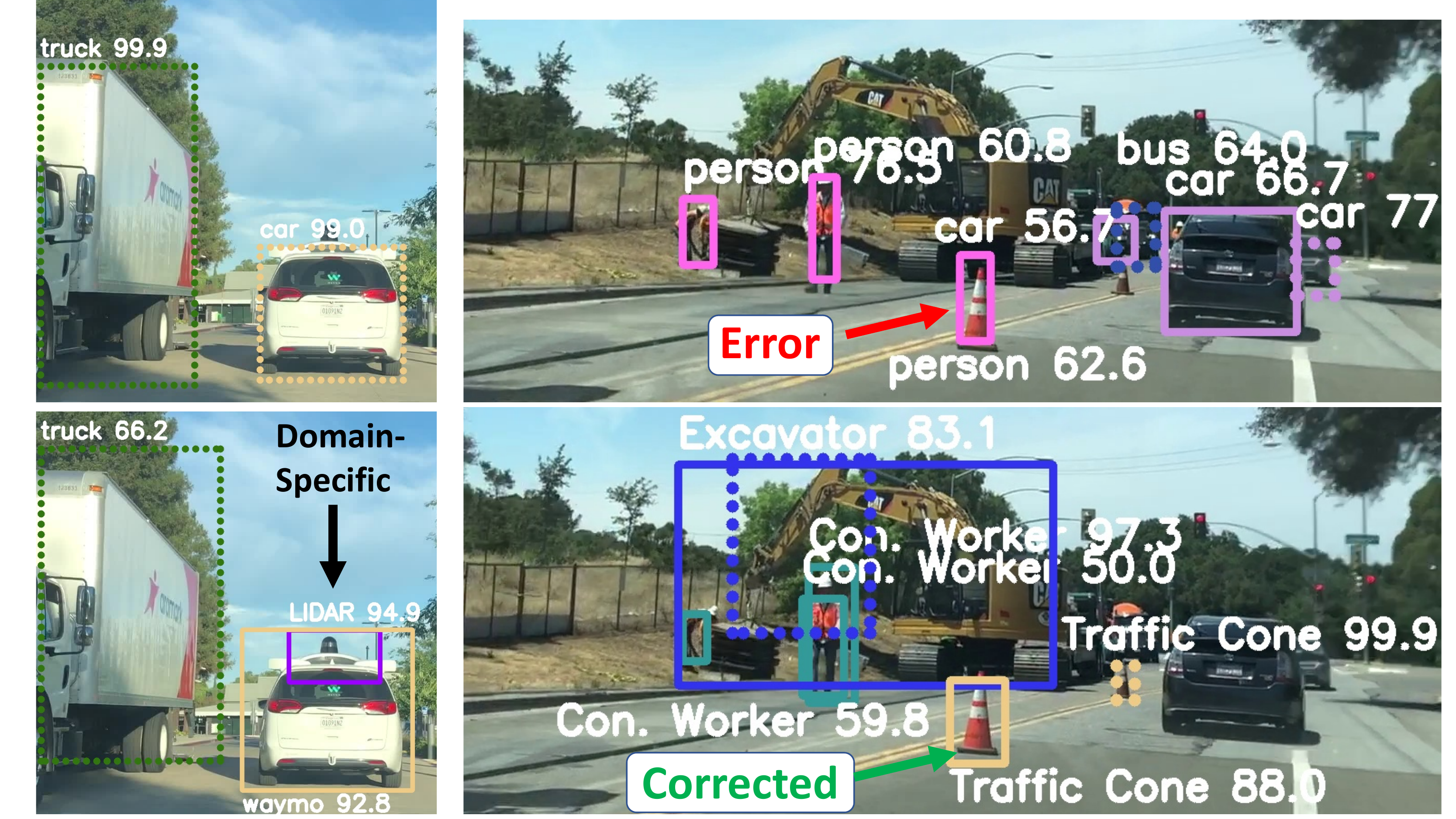}
    \caption{\textbf{\name Re-Trained Vision Models.}
    From only a minimal set of training examples, \name adapts default vision models trained on the standard COCO dataset \cite{coco} (top images) to better, domain-specific models (bottom images). We flexibly fine-tune mobile-optimized models, such as MobileNet 2, as well as compute-intensive, but more accurate models such as \maskrcnn \cite{ren2015faster}.} 
%Objects detected \textit{only} by R-CNN are shown in dashed lines.} 

	\label{fig:contrast_retrain}
\end{figure}

% NEW STUFF MARCO ADDED
In general, the problem of intelligently sampling useful or anomalous training data
is of extremely wide scope, since field robotic data
contains both \textit{known weakpoints} of a model as well as \textit{unknown weakpoints} that have yet to even be discovered.
A concrete example of a known weakpoint is shown in Fig. \ref{fig:contrast_retrain}, 
where a fleet operator identifies that construction sites are common errors of a perception model and wishes to collect more of such images, perhaps because they are insufficiently represented in a training dataset.
In contrast, unknown weakpoints could be rare visual concepts that are not even anticipated by a roboticist. 

To provide a focused contribution, this paper addresses the problem of how robotic fleets can source more training data for \textit{known weakpoints} to maximize model accuracy, but with minimal end-to-end systems costs. Though of tractable scope, this problem
is still extremely challenging since a robot must harvest rare, task-relevant training examples
from growing volumes of robotic sensory data. 
A principal benefit of addressing \textit{known} model weakpoints is 
that the cloud can directly use knowledge of task-relevant image classes, which require more training data,
to guide and improve a robot's sampling process.
Our key technical insight is to keep robot sampling logic simple, but to leverage the power of the cloud
to continually re-train models and annotate uncertain images with a human-in-the-loop, thereby
using cloud feedback to direct robotic sampling. 
Moreover, we discuss how insights
from \name can be extended to collecting training data for unknown weakpoints in future work. 

\tu{Literature Review:}
\name is closely aligned with \textit{cloud robotics}
\cite{goldbergwebsite,kuffner2010cloud,kehoe2015survey}
, where robots send video or LIDAR to the cloud to query compute-intensive models or databases when they are uncertain. Robots can face significant network congestion while
transmitting high-datarate sensory streams to the cloud, leading to work that selectively offloads only uncertain samples \cite{chinchali2019RSS} or uses low-latency ``fog computing'' nodes for object manipulation and tele-operation \cite{tanwani2019fog,tian2018fog}. 
As opposed to focusing solely on network latency for real-time control, \name addresses overall system cost (from network transfer to human annotation time) and instead focuses on the open problem of continual learning
from cleverly-sampled batches of field data.

\RSS{Our work is inspired by traditional active learning \cite{tong2001active, settles2009active, cohn1996active, gal2017deep}, novelty detection \cite{perera2019deep,bodesheim2013kernel}, and few-shot learning \cite{zhu2019generalized,wang2019tafe, sanakoyeu2019divide}, but differs significantly since we use novel compute-efficient algorithms, distributed between the robot and cloud, that are scalable to run on even resource-constrained robots. 
Many active learning algorithms \cite{tong2001active, settles2009active, cohn1996active}
use compute-intensive \textit{acquisition functions} to decide which limited samples a learner should label to maximize model accuracy, which often involve Monte Carlo Sampling to estimate uncertainty over a new image. Indeed, the most similar work to ours \cite{gal2017deep} uses Bayesian Deep Neural Networks (DNNs), which have a \textit{distribution} over model parameters, to take \textit{several} stochastic forward passes of a neural network to empirically quantify uncertainty and decide whether an image should be labeled. Such methods are infeasible for resource-constrained robots, which might not have the time to take several forward passes of a DNN for a single image or capability to efficiently run a distribution over model parameters. Indeed, we show in Section \ref{sec:results} how our sampler uses a 
hardware DNN accelerator, specifically the Google Edge Tensor Processing Unit (TPU), to evaluate an on-board vision DNN, with \textit{one fixed set} of pre-compiled model parameters, in $<16$ms and decides to sample an image in only $\approx0.3$ms!
More broadly, our work differs since we (1) address sensory streams (as opposed to a pool of unlabeled data), (2) use a compute-efficient model to guide sampling, (3) direct sampling with task-relevant cloud feedback, and (4) also address storage, network, and computing costs.} 
%Our work further differs from traditional novelty detection \cite{perera2019deep,bodesheim2013kernel} and few-shot learning \cite{zhu2019generalized,wang2019tafe, sanakoyeu2019divide}, which identify new or outlier visual classes from minimal examples, but often with compute-intensive methods.}

\tu{Statement of Contributions:}
In light of prior work, our contributions are four-fold. First, we collect months of novel video footage to quantify the systems bottlenecks of processing growing sensory data. 
Second, we empirically show that large improvements in perception model accuracy can be realized by automatically sampling task-relevant training examples from vast sensory streams.
Third, we design a sampler that intelligently delegates compute-intensive tasks to the cloud, such as adaptively tuning key parameters of robot sampling policies via cloud feedback.
Finally, we show that our perception models run efficiently on the state-of-the-art 
Google Edge TPU \cite{edgeTPUwebsite} for resource-constrained robots.

\tu{Organization:}
This paper is organized as follows. Section \ref{sec:motivation} motivates the \textit{Robot Sensory Sampling Problem}
by quantifying our system costs for collecting three novel perception datasets. 
Then, Section \ref{sec:prob_statement} formulates a general sampling problem and introduces the \name system architecture (Fig. \ref{fig:sys_architecture}). Next, Sections \ref{sec:sampler_algorithm} and \ref{sec:results} 
quantify \name's accuracy improvements for perception models, key reductions in system bottlenecks, and performance on the Edge TPU \cite{edgeTPUwebsite}. 
Finally, we conclude with future applications in Section \ref{sec:discuss}. 

% RETRAIN IMAGE
%%%%%%%%%%%%%%%%%%%%

\section{Motivation}
\label{sec:motivation}
We now describe the costs and benefits of
continual learning in the cloud by collecting and processing three novel perception datasets.

\subsection{Costs and Benefits of Continual Learning in the Cloud}
\label{subsec:benefit_cloud_retrain}

\paragraph*{Benefits} A prime benefit of continual learning in the cloud is to correct weak points of existing models or specialize them to new domains.
As shown in Fig. \ref{fig:contrast_retrain} (right, top), standard, publicly-available vision deep neural networks (DNNs) pre-trained on benchmark datasets like MS-COCO \cite{coco} can misclassify traffic cones as humans but also miss the large \textit{excavator} vehicle, which neither looks like a standard truck nor a car. However, as shown in Fig. \ref{fig:contrast_retrain} (right, bottom), a specialized model re-trained in the cloud using automatically sampled, \textit{task-relevant} field data can correct such errors. More generally, field robots that ``crowd-source'' interesting or anomalous training data could automatically update high-definition (HD) road maps.

\paragraph*{Costs}
\RSS{The above benefits of cloud re-training, however, often come with understated costs, which we now quantify. Table \ref{fig:system_cost_table} estimates monthly systems costs for a small fleet of 10 self-driving cars, each of which is assumed to capture 4 hours of driving data per day. In the first scenario, called ``Dashcam-ours'', we assume a single car has only one dashcam capturing H.264 compressed video, as in our collected dataset. In the second scenario, called ``Multi-Sensor'', we base our calculations off of an Intel report \cite{intel} that estimates each SDV will generate 4 TB of video and LIDAR data in just 1.5 hours.}

\RSS{To be conservative, we assumed all field data is uploaded over fast 10 Gbps ethernet, which might be realistic for SDVs, but is much slower for  
    network-limited robots. Most importantly, we assume \textit{only 1\% of image data} is interesting and needs to be annotated, specifically by the state-of-the-art Google Data Labeling service \cite{cloud_data_label}, which has standard rates for object detection bounding boxes and (costlier) segmentation masks (referred to as ``B. Box'' and ``Mask'' in Table \ref{fig:system_cost_table}).
Further, to roughly estimate annotation time, we hand-labeled 1000 bounding boxes in 1.4 hours. Since we did not hand-label nor train perception models for semantic segmentation, the time estimates are marked as ``NA''. 
Though conservative, our estimates, which are further detailed in supplementary information, show significant costs, especially for annotation. 
Indeed, we confirmed the sampling problem's relevance by talking with robotics startups that face high systems costs.}

% NON MULTI-COLUMN VERSION
\begin{table}[t]
    \centering
	\begin{tabular}{|l|l|l|l|l|}
		\hline
		\multirow{2}{*}{ \textbf{Bottleneck}} &
       \multicolumn{2}{c}{\textbf{Multi-Sensor}} &
       \multicolumn{2}{c|}{\textbf{Dashcam-ours}} \\
	     &  \textbf{B. Box} & \textbf{Mask} & \textbf{B. Box} & \textbf{Mask}\\ 
		\hline
		\cellcolor{Gray!20} Storage & \cellcolor{CornflowerBlue!20} \$31,200 & \cellcolor{YellowGreen!20} \$31,200  & \cellcolor{CornflowerBlue!20} \$172.50 & \cellcolor{YellowGreen!20} \$172.50 \\
		\hline
        \cellcolor{Gray!20} Transfer Time (hr) & \cellcolor{CornflowerBlue!20} 8.88  & \cellcolor{YellowGreen!20} 8.88 & \cellcolor{CornflowerBlue!20} 0.05 & \cellcolor{YellowGreen!20} 0.05 \\
		\hline
		\cellcolor{Gray!20} Annotation Cost & \cellcolor{CornflowerBlue!20} \$95,256 & \cellcolor{YellowGreen!20} \$1,652,400 & \cellcolor{CornflowerBlue!20} \$15,876 & \cellcolor{YellowGreen!20} \$275,400 \\
		\hline
        \cellcolor{Gray!20} Annot. Time (days) & 22 \cellcolor{CornflowerBlue!20} & NA \cellcolor{YellowGreen!20} & 18 \cellcolor{CornflowerBlue!20} & NA \cellcolor{YellowGreen!20} \\
		\hline
	\end{tabular}
	\caption{\textbf{Systems Costs:}
    Monthly storage/annotation cost and daily network upload time are significant for self-driving car data even if we annotate \textit{1\% of video frames}.}
	\label{fig:system_cost_table}
\end{table}

\subsection{Novel Datasets we Collected}
\label{subsec:sampler_need}
To quantify systems costs first-hand, we collected over three months of field data, comprising over \SC{10 GB} of compressed video clips from a dashboard iPhone 6 camera (``dashcam''). \SC{While modest in size, it contains \SC{346,788} frames of video, which is challenging to automatically mine for training examples.} The following datasets, as well as re-trained vision models, are available at \RSS{\url{https://sites.google.com/view/harvestnet}}:

\noindent \tu{1. Identifying Construction Sites:}
Our three months of construction footage shows safety-critical scenarios and model drift, including roads that were first excavated by tractors (Fig. \ref{fig:contrast_retrain}), filled with tar, and rolled back to normal by ``compactor'' machines and human workers. 

\noindent \tu{2. Identifying Self-driving Vehicles (SDVs):}
We captured rare footage of SDVs being tested on public roads over 3 months, predominantly of the Waymo van (Fig. \ref{fig:contrast_retrain}). 
This example is a proof-of-concept of \name's ability to sift rare, task-relevant training samples from large volumes of regular car footage, and is easily generalizable to other image classes.

\noindent \tu{3. Streaming Face Recognition (FaceNet):}
To test \name's robustness to domains outside of road scenes, 
we use a video dataset of streaming faces of consenting volunteers from a prior robotics paper \cite{chinchali2019RSS}.
We expressly do not advocate privacy-infringing surveillance missions, 
and simply test a search-and-rescue scenario where a sampler must use face 
embeddings, such as from the de-facto FaceNet \cite{schroff2015facenet,openface} DNN, to sift faces of interest from large volumes of video. 
Our results on all three datasets, henceforth referred to as Waymo, Construction, and FaceNet, are shown in Section \ref{sec:results}.

\section{Problem Statement}
\label{sec:prob_statement}
% SINGLE IMAGES
\begin{figure}[h]
	\centering
	\includegraphics[width=1.0\columnwidth]{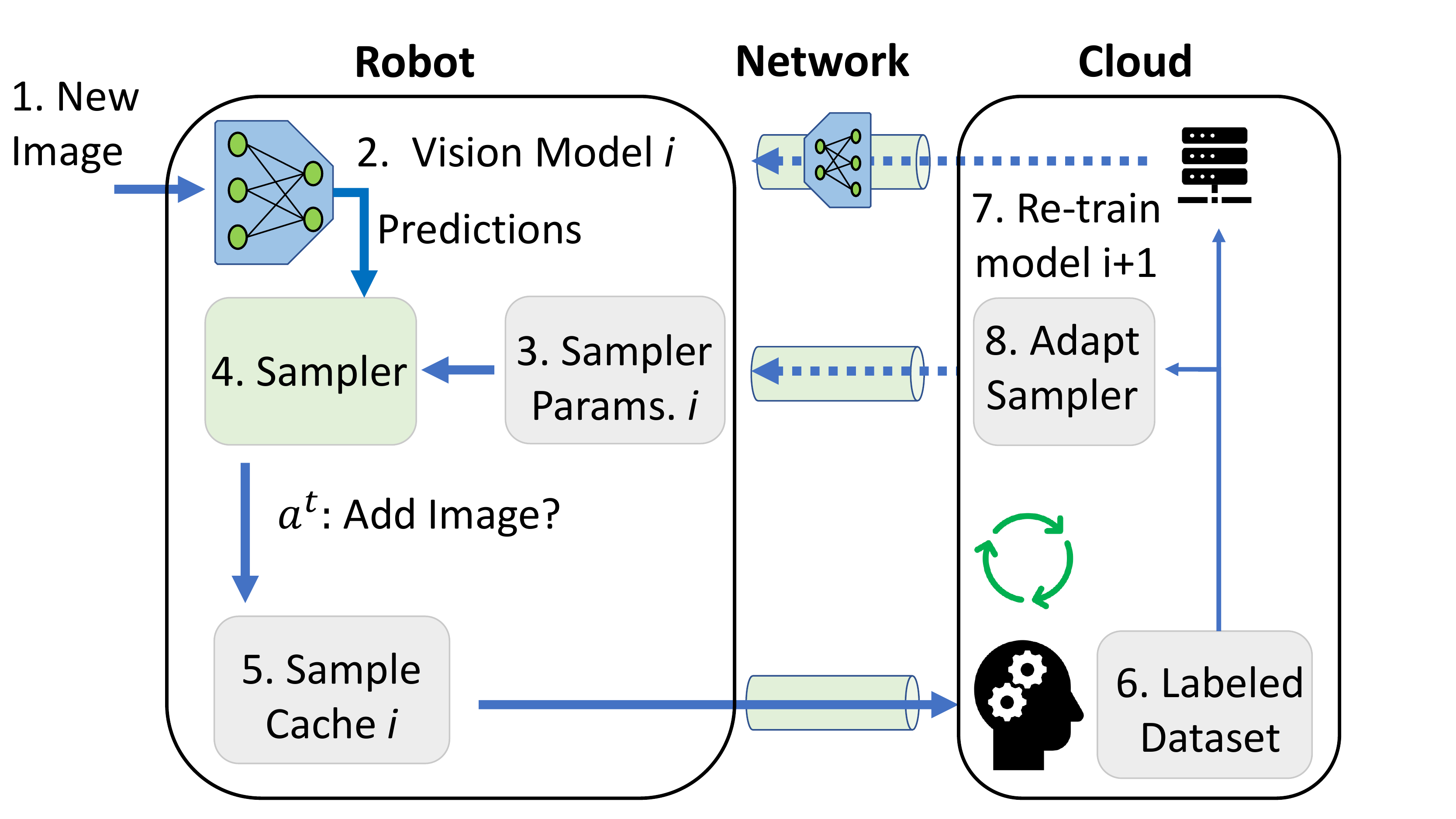}
    \caption{\textbf{System Architecture:} Our core contribution is a robot sensory sampler in green, 
    which uses predictions from an on-board vision DNN and cloud-selected image targets.}
	\label{fig:sys_architecture}
\end{figure}

We now formulate a general \textit{Robot Sensory Sampling Problem}
that is applicable to diverse field robots.
Examples include self-driving cars, low-power drones, and even mining robots
which might only have network access when they surface. As a common theme, these robots must \textit{prioritize} sensory samples due to limits in some combination of on-board storage, network connectivity, human annotation, and/or cloud-computing time that preclude na\"ively using all field data. We now define general abstractions for the sampling problem.

%Examples range from Mars Rovers that can only query scientists and download models over a slow space link to mining robots which might only have network access when they surface. As a common theme, these robots must \textit{prioritize} sensory samples due to limits in on-board or cloud storage, network connectivity, human annotation, and/or cloud-computing time that preclude na\"ively using all field data. We now define general abstractions for the sampling problem.

\noindent        \tu{1. Sensory Input:} 
A robot measures a stream of sensory inputs $\{x^{t,i}\}_{t=0}^{T}$, where each $x^{t,i}$ is a new image or LIDAR point cloud, on a day or learning ``round'' $i$. \RSS{At the end of a day's horizon of $T$ steps, it can upload samples to the cloud for downstream annotation and model learning. In practice, time horizon $T$ can be the duration of a day's field test when a subterranean robot surfaces and connects to a network.}
Since the robot is caching images for model-learning, as opposed to real-time control, it can sub-sample the sensory input at a sustainable rate, such as $\Delta t = 30s$, by skipping frames.
    
\noindent       \tu{2. Robot Perception Models:} Vision DNN $\fDNNrobot$ maps an image $x^t$ to a prediction $\yhat^t$ (e.g. image class or object detections) and associated confidence $\yconf^t$, typically from softmax layers in the DNN. Further, the DNN also provides embeddings $\emb^t$, which can be used to compare the \SC{similarity between images} as shown in Fig. \ref{fig:diminishing_returns_plot} (right). 
The DNN outputs a vector of predictions, confidences, and embeddings, defined as: 
    \begin{small}
    \begin{align}
        \label{eq:perception_DNN}
        \boldy^t &= [\yhat^t, \yconf^t, \emb^t] = \fDNNrobot(x^t).
    \end{align}
    \end{small}
    \RSS{We denote the perception DNN learned after round $i$ as $\fDNNrobot^{i}$, where $\fDNNrobot^{0}$ is the initial DNN the robot is equipped with before re-training}. 
    Our goal is to make $\fDNNrobot^{i}$ as accurate as possible for increasing $i$ with the fewest samples sent to the cloud. Fixed model $\fDNNrobot^{i}$ is not re-trained \textit{within} a round $i$ of sample collection \RSS{since the samples must first be annotated}. 
    
\noindent        \tu{3. Target Image Set:}
As stated in the introduction, the scope of this paper is to collect more examples of \textit{known weakpoints} of a model, such as the mis-labeled construction sites in Fig. \ref{fig:contrast_retrain}. In practice, these weakpoints, which we refer to as task-relevant image classes, can be identified by a fleet operator who might nightly review model performance on historical data.
The set of task-relevant image classes, such as traffic cones, excavators, and LIDAR units (Fig. \ref{fig:contrast_retrain}), are denoted by $\targetclassset$. 
Importantly, we assume the cloud has a few initial examples of images belonging to $\targetclassset$, denoted as $\targetimageset$. This is a practical assumption, since $\targetimageset$ can be obtained from a few examples of mis-labeled field data, like Fig. \ref{fig:contrast_retrain}, that initially prompted the problem, or even collated from Google Images. Equally importantly, the cloud dataset can already have sufficient training data for well-understood scenes from previous field tests or public datasets like ImageNet \cite{deng2009imagenet}, to ensure model performance does not degrade on such data.

\noindent        \tu{4. Sampling Policy:}
\RSS{Our goal is to develop a lightweight sampling policy $\pisample$ (green unit in Fig. \ref{fig:sys_architecture}) that decides whether or not to store a new image $x^t$ based on predictions from the on-board perception DNN $\boldy^t = \fDNNrobot^i(x^t)$ and the desired target-set $\targetclassset$ of images to look for}:  
    \begin{small}
    \begin{align}
        \label{eq:pi_sample}
        a^t = \pisample(x^t, \boldy^t, \targetclassset, \thetasampler^{i}),  
    \end{align}
    \end{small}
    \RSS{where $\thetasampler^i$ are sampler hyper-parameters that, in general, can be adapted per round $i$}. Here, $a^t = 1$ indicates storing the image to later send to the cloud and $a^t = 0$ indicates dropping it, such as for common, \RSS{well-understood} scenes.
    \RSS{Ideally, to work on resource-constrained robots as well as not interfere with real-time decision-making, $\pisample$'s logic, including use of hyper-parameters $\thetasampler^{i}$,  should be compute-efficient.} 
%An efficient solution, which works on embedded platforms such as the Google Edge TPU, is presented in Sec. \ref{sec:sampler_algorithm}}.

\noindent        \tu{5. Limited Robot Sample Cache:} 
In round $i$, the robot saves 
sampled images in a small cache denoted by $\Cache^i = \{x^\tau\}_{\tau = 0}^{\Cachesize}$, where $0 \le \tau \le T$. These images, limited by a size $\Cachesize \ll T$, are small enough to store on-board a robot and will be uploaded to the cloud at the end of round $i$. 
    
\noindent        \tu{6. Annotated Cloud Dataset:} When 
samples in $\Cache^i$ are uploaded to the cloud, a human, or trusted machine, provides ground-truth annotations $\yground^t$ for the selected images, which could be different from the robot's \textit{predictions} $\yhat^t$. In the cloud, we store a dataset $\dataset^i = \{x^{\tau}, y^{\tau}\}$ that grows with rounds $i$ and comprises images $x^\tau$ and ground-truth labels $y^\tau$: 
    
    \begin{small}
    \begin{align}
        \label{eq:dataset_update}
        \dataset^i = \dataset^{i-1} ~\cup~ \{(x^\tau, \yground^\tau) ~, ~\forall x^\tau \in \Cache^i\}.
    \end{align}
    \end{small}
    Initial dataset $\dataset^0$ has very few examples of the target images that we care for, such as self-driving cars or road disruptions. However, initial dataset $\dataset^0$ can be seeded with a large set of ``common'' training examples, such as from ImageNet \cite{deng2009imagenet}. In the cloud, we split full dataset $\dataset^i$ into train, validation, and test components denoted by $\dtrain^i$, $\dval^i$, and $\dtest^i$. 
    
\noindent        \tu{7. Independent Test Dataset $\dtestfinal$:} 
\RSS{Model accuracy is always reported on an \textit{independent, held-out} test set denoted $\dtestfinal$. To accurately assess the efficacy of a sampler in accruing \textit{task-relevant} data that yields steady performance gains on the target classes, 
$\dtestfinal$ should have sufficient representation, or even be biased towards, target classes. The rest of the images can come from cloud datasets like ImageNet, if we wish to ensure performance does not degrade on these classes.
In the extreme case where we only desire a specialized model, such as for road scene classes only, $\dtestfinal$ can also fully contain only target class images. 

In practice, $\dtestfinal$ can be sourced from designated robots in a fleet tasked with collecting \textit{only test} data. If special test robots are scarce, $\dtestfinal$ can be accumulated from a random subset of data annotated by a human per round $i$. In both scenarios, we ensure test data is independently verified by annotators and never used for training.}

\noindent        \tu{8. Model Re-training and Evaluation:}
Once the new annotated cache $\Cache^i$ is added to the dataset $\dataset^i$, model re-training function $\gretrain$ takes current robot model $\fDNNrobot^i$ and the updated annotated dataset $\dataset^{i+1}$ and outputs an updated model $\fDNNrobot^{i+1}$. 
    The training objective is to minimize loss function $\mathcal{L}(\yhat^t, \yground^t)$ penalizing prediction errors, such as the standard cross-entropy loss for image classification or mean average precision (mAP) for object detection \cite{coco_detection_metrics}. Then, evaluation function $\gevaluate$ quantifies the accuracy improvements of newly trained model $\fDNNrobot^{i+1}$ given a validation dataset, $\dval^i$, or test set $\dtestfinal$.
    The model re-train and evaluation functions are denoted by:
    \begin{small}
    \begin{align}
        \label{eq:model_retrain}
        \fDNNrobot^{i+1} \leftarrow \gretrain(\fDNNrobot^i, \dtrain^{i+1}) \\
        \label{eq:model_evaluation}
        \mathcal{L}^{i}= \gevaluate(\fDNNrobot^i, \dtestfinal).
    \end{align}
    \end{small}
    % keep 
    Finally, new model $\fDNNrobot^{i+1}$ is downloaded back to the robot over a network link.    \RSS{Vision DNN models are typically less than a few GB, which are very small, in contrast to raw sensory data, to download daily even over bandwidth-limited networks.}

\subsection{The Robot Sensory Sampling Problem (Fig. \ref{fig:model_retrain})}
\label{subsec:prob_statement}
We now formally define the abstract sampling problem.  
\begin{problem}[Robot Sensory Sampling Problem]
\label{problem:rss_problem}
\RSS{Given a target set of image classes $\targetclassset$,
initial cloud dataset $\dataset^{0}$, and initial robot model $\fDNNrobot^{0}$, 
find the best sampling policy $\pisample^{*}$ that minimizes full held-out test set loss $\dtestfinal$ 
for the final robot DNN $\fDNNrobot^{\roundmax}$ trained over $\roundmax$ days:} \\

\begin{small}
\begin{align}
\label{eq:rss_problem}
\pisample^*\in \argmin_{\pisample} \mathcal{L}(\fDNNrobot^{\roundmax}, \dtestfinal), 
\end{align}
\end{small}
\end{problem}
\RSS{subject to the constraint that a robot can only sample $\Cachesize$ images daily.}

\tu{Sampling Problem Complexity:}
\RSS{Deriving an optimal sampling policy that solves Problem \ref{problem:rss_problem} is extremely challenging for several reasons. First, it is hard to analytically model the sequence of high-dimensional sensory inputs $\{x^{t,i}\}_{t=0}^{T}$ a robot can possibly measure on a day $i$. Second, even if such an observation model exists, it is extremely challenging to 
decide which limited subset of $\Cachesize$ samples, chosen from $T$ possibilities, will lead to the best model performance of a complex DNN evaluated on a final test-set \textit{at the end} of several rounds $\roundmax$. 
Indeed, even for a small drone that measures video \textit{for only 15 minutes} at a slow frame-rate of 5 frames per second with a cache of $\Cachesize = 2000$ daily images, there is a combinatorial explosion of $ \binom{T}{\Cachesize} = 4.21 \times 10^{1340}$ images to choose from for downstream model learning!}
%Indeed, even for a small drone that measures video at 30 frames per second (FPS) \textit{subsampled evenly at 5 FPS} for only 15 minutes with a cache of $\Cachesize = 2000$ daily images, there is a combinatorial explosion of $ \binom{T}{\Cachesize} = 4.21 \times 10^{1340}$ images to choose from for downstream model learning!}

\tu{Wide Practicality of the Sensory Sampling Problem:}
\RSS{Despite its complexity, the sampling problem is built on simple, but general, abstractions of (a) a robot perception model, (b) limited on-board storage, (c) occasional network access to upload samples, and (d) centralized annotation and compute resources. These abstractions 
apply to a wide variety of robots where the \textit{rate} of sensory measurements outpaces that of human annotation and downstream cloud processing.}

\section{The \name Intelligent Sampler} 
\label{sec:sampler_algorithm}
\RSS{While obtaining a provably optimal solution to Problem \ref{problem:rss_problem}
is infeasible, one can envision a myriad of possibly well-performing heuristics.
To ensure \name applies to a wide diversity of robots, we enforce a practical requirement - the sampler should be sufficiently compute-efficient to run on even a resource-constrained robot, 
equipped with, for example, a Raspberry Pi and Google Edge TPU as shown in Fig. \ref{fig:RPI_sampler}. A minimalistic on-board sampler that does not interfere with real-time control
is extremely useful for larger robots as well, such as SDVs. 
We now describe three empirical insights, 
evidenced in Fig. \ref{fig:diminishing_returns_plot}, that constrain the sampler design space and directly lead to \name.} 

\subsection{Empirical Insights Behind the \name Sampler}
\label{subsec:empirical_insights}
%%%%%%%%%%%%%%%%%%%%%%%%%%%%%%%%%%%
\begin{figure}[t]
    \centering
    \includegraphics[width=1.0\columnwidth]{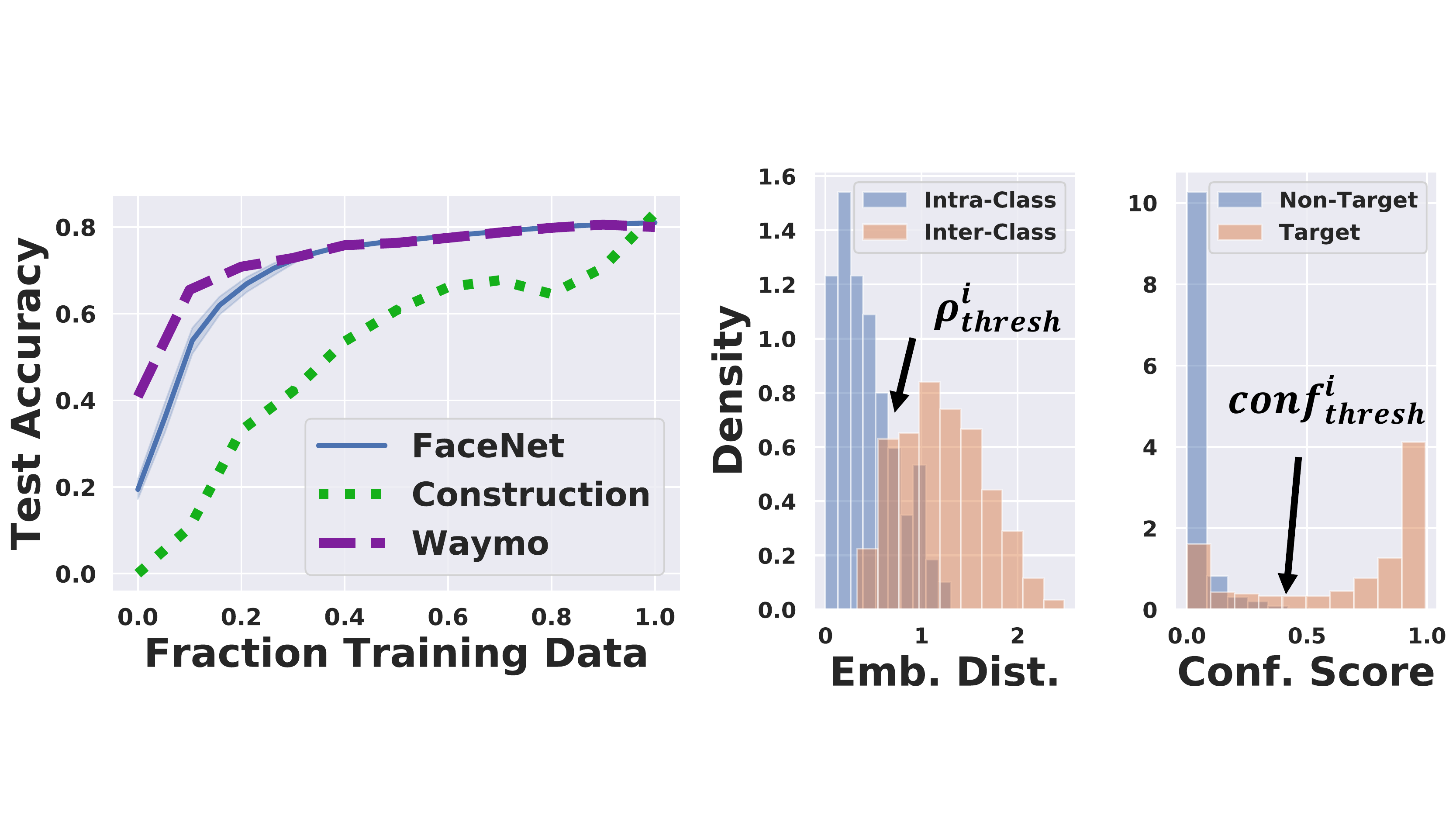}
    \caption{\textbf{Empirical insights guiding \name}: (Left) 
        For relatively static targets, classification test accuracy saturates with more images used to fine-tune a vision DNN. 
    (Right) We use either embedding distances or softmax confidence scores to signal if a new image is likely a target of interest.}
    \label{fig:diminishing_returns_plot}
\end{figure}

%%%%%%%%%%%%%%%%%%%%%%%%%%%%%%%%%%%

\noindent \textbf{Insight 1:} \textit{Use transfer learning for high model accuracy with few sampled images.\\}
Many pre-trained CNNs' initial network layers serve as general-purpose feature extractors that yield finite-dimension \textit{embeddings} \cite{sharif2014cnn,yosinski2014transferable,CS231N_transfer_learning}. Embeddings can be used to 
    \textit{fine-tune} final linear classification or object detection layers to recognize new classes of interest, such as autonomous cars or construction sites, that a CNN was not initially trained upon.

    \name exploits fine-tuning by our empirical observation that only a limited set of images are necessary to adapt a pre-trained vision DNN to new domains, especially if the new domains' data does not change much with time.
Fig. \ref{fig:diminishing_returns_plot} (left) shows the full dataset test accuracy of our three separate datasets (y-axis) versus the percentage of total training data provided to the learning model (x-axis). Clearly, for the relatively \textit{static} Waymo and FaceNet domains, the final model accuracy saturates as more training images are acquired.

\RSS{In contrast, for the \textit{dynamic} task of identifying construction sites, the accuracy steadily increases as more data is acquired, even after months of data. This is because construction sites are constantly changing, with new tractors and signs being updated daily. Indeed,
\name can be used for \textit{continual} re-training, allowing robots to update HD maps in the cloud.} 
%with temporary road-blocks or pot-holes in a cloud database.} 
%in the cloud in a data-efficient manner.} 
%to improve accuracy by constantly re-training on dynamic targets, such as temporary road-blocks or pot-holes which can be used to update HD maps for robots.

\noindent \textbf{Insight 2:} \textit{Filter task-relevant target images using DNN embeddings and softmax confidences. \\}

\paragraph*{Embedding distances} 
Many, but not all, CNNs have the property that the $L_2$ distance between embeddings of two images is a proxy for image similarity. 
In particular, FaceNet DNN embeddings typically have a lower embedding-distance for two pictures of the \textit{same person} than unrelated faces, since the DNN is optimized with a triplet loss \cite{openface,schroff2015facenet}. We test this property with our FaceNet data in Fig. \ref{fig:diminishing_returns_plot} (right), allowing \name to exploit embedding distance as a valuable indication of whether a new image is of the same desired class as training data we wish to acquire.

\paragraph*{Softmax confidence thresholds} 
Softmax confidence scores yield a probability distribution over CNN class predictions. 
We empirically note that, while the softmax probabilities are not necessarily a \textit{formal} probability measure, images that are indeed truly a class (``true positives'') have a higher softmax confidence element for that class of interest compared to ``true negatives''. 
For example, as shown in Fig. \ref{fig:diminishing_returns_plot} (right),
images that are indeed of Waymo cars (red) have a higher confidence score for the Waymo class output compared to negatives (in blue), showing that confidence scores above a separating threshold are a useful signal to store a sample of interest. 

Crucially, embeddings and softmax confidence scores are readily produced simply by running a DNN forward-pass for inference. Thus, when a robot's perception model is already being run for a high-level task, such as motion planning, the sampler obtains the embeddings and confidence scores ``for free''. 
Then, these two metrics can be used in a lightweight, data-driven filter to discern if a new image is likely a target of interest for caching. We further quantify in Sec. \ref{subsec:TPU_performance} how we can run \name's models on the Edge TPU \cite{edgeTPUwebsite}.

\begin{figure}[t]
    \centering
    \includegraphics[width=1.0\columnwidth]{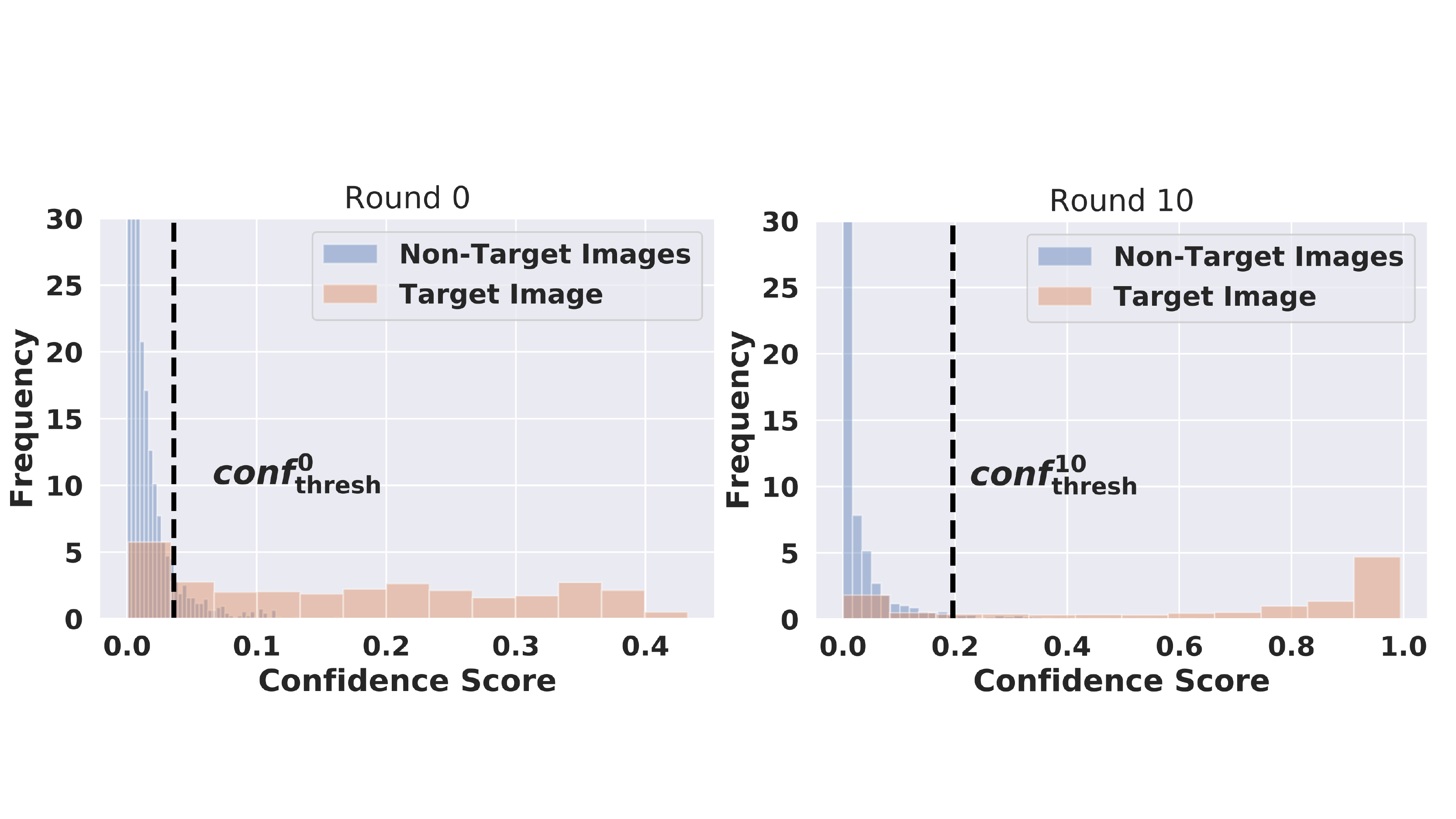} 
    \caption{\RSS{\textbf{Role of Cloud Feedback}: After a human annotates more samples on each round $i$, the softmax confidence distribution difference between target (red) and irrelevant (blue) images becomes sharper, which \name uses to improve sampling by adapting the confidence threshold $\yconfthresh^{i}$.}}
    \label{fig:adapt_conf_threshold}
\end{figure}

\noindent \textbf{Insight 3:} \textit{Use prioritized human feedback in the cloud to refine the target image filter. \\}
\RSS{The final insight behind \name, shown in Fig. \ref{fig:adapt_conf_threshold},
is that the DNN softmax confidence distributions between target images (red) and irrelevant images (blue) become more differentiated as more human-annotated images are added to the cloud dataset at every round $i$. Crucially, this allows the cloud to better signal to the robot's sampler how to identify task-relevant target images from background noise, thereby directly building on insight 2.}
A broader insight behind \name is that we can delegate compute-intensive tasks of training to the cloud, but provide simple feedback to a robot to improve its target image filter.

%%%%%%%%%%%%%%%%%%%%%%%%%%%%%%%%%%%

\subsection{Practical application of the \name insights}
\label{subsec:application_insights}
\RSS{We now formalize how, together, insights 1-3 directly lead to our novel \name sampler, introduced in Algorithm \ref{alg:harvestnet}. 
Importantly, \name is markedly distinct from today's active learning and data selection algorithms, due to its use of a compute-efficient robot model and adaptive feedback from cloud computing resources.} 

\noindent \paragraph*{\textbf{Application of Insight 1 (Fine-Tuning)}}
\RSS{Insight 1, which governs how to fine-tune a DNN with minimal images (Fig. \ref{fig:diminishing_returns_plot}, left), is addressed by using pre-trained feature extraction layers for initial robot model $\fDNNrobot^{0}$, such as from publicly available models, and only re-training final layers with transfer learning for new target image classes. Further, we limit the sampled cache size by $\Cachesize$}.

\noindent \paragraph*{\textbf{Application of Insight 2 (Simple Target Image Filter)}}
\RSS{Insight 2, which governs how to use DNN softmax confidences or embedding distances to build a compute-efficient on-board target image filter (Fig. \ref{fig:diminishing_returns_plot}, right), is now formalized in two variants of sampling policies $\pisample$ (Eq. \ref{eq:pi_sample}).}

\tu{Variant 1. Embedding Distance Based Sampling:}
\RSS{Two pictures of the \textit{same person} $x^{t_1}$ and $x^{t_2}$ typically have a lower embedding-distance $\rho(x^{t_1}, x^{t_2}) = \left \Vert \emb^{t_1} - \emb^{t_2} \right \Vert ^{2}_{2}$ than unrelated faces \cite{schroff2015facenet}.
Using this insight, we compare the embedding distance between a robot's new sensory input $x^t$ to each of the target images in $\targetimageset$, and cache the sample if the median distance is below an empirical threshold $\rhothresh^{i}$, indicating similarity to a face we want to learn. 
Embedding distance threshold $\rhothresh^{i}$ serves as the sampler hyper-parameter $\thetasampler^{i} = \{\rhothresh^{i}\}$ in Eqn. \ref{eq:pi_sample}, yielding:}

\begin{small}
    \begin{align*}
        a^t & = \pisample(x^t, \boldy^t, \targetimageset, \thetasampler^{i}) \\ 
            &= \mathbb{1} \left(\textsc{Median} \left\{\rho(x^t, x^{\mathrm{targ.}}),~ \forall x^{\mathrm{targ.}} \in \mathrm{TargImSet} \right\} < \rhothresh^{i} \right). \\  
    \end{align*}
\end{small}
\RSS{In this variant of the sampling policy, the cloud sends the $\targetimageset$ of images and their embeddings to the robot to compare with the embedding of a new image $x^t$. In practice, 
    it is trivial to store a limited set of target images on-board, such as 18 images which sufficed in our experiments.}

\tu{Variant 2. Classification Confidence Based Sampling: }
\RSS{In cases where embedding distances are not optimized for similarity between two input images, we can directly use confidence scores $\yconf^t$ to determine if a new image is likely to be a useful representative of the desired target. We rely on the fact that initial robot DNN $\fDNNrobot^{0}$, as well as successive versions $\fDNNrobot^{i}$, are trained with a few examples of target images and can predict their target class $\targetclass$ in the DNN model architecture. Then, given a new image $x^t$, we see if the confidence element \textit{for a target class}, denoted by $\yconf[\targetclassset]$, is above an empirically determined threshold $\yconfthresh^{i}$  (Fig. \ref{fig:adapt_conf_threshold}), which serves as the sampler hyper-parameter $\thetasampler^{i} = \{\yconfthresh^{i}\}$ in Eqn. \ref{eq:pi_sample}. 
Thus, the sampling policy becomes:}
\begin{small}
    \begin{align*}
        a^t & = \pisample(x^t, \boldy^t, \targetclassset, \thetasampler^{i}) \\ 
            & = \mathbb{1} \left(\yconf^{t}[\targetclass] > \yconfthresh^{i} \right). \\  
    \end{align*}
\end{small}

\RSS{Unlike the embedding-distance variant of the sampling policy, the softmax confidence version simply needs to know the desired target classes $\targetclassset$ to harvest more images of, which does not require storing any target images $\targetimageset$ on-board. As measured in Sec. \ref{sec:results}, it is trivial to do a comparison of the output softmax confidences of a DNN with a fixed threshold efficiently on embedded-devices.}

\noindent \paragraph*{\textbf{Application of Insight 3 (Cloud Feedback)}}
\RSS{Our key insight, depicted in Fig. \ref{fig:adapt_conf_threshold}, is to use the new cloud dataset and improved model at the \textit{end of round $i$}, $\dataset^{i+1}$ and $\fDNNrobot^{i+1}$, to best adapt the sampler hyper-parameters $\thetasampler^{i+1} = \{\yconfthresh^{i+1}, \rhothresh^{i+1}\}$ for the next round. This is achieved by the $\updatethreshold$ function:}

\begin{small}
\begin{align}
    \label{eq:update_threshold}
    \thetasampler^{i+1} \leftarrow \updatethreshold(\dataset^{i+1}, \fDNNrobot^{i+1}).
\end{align}
\end{small}

\RSS{In practice, $\updatethreshold$ can be implemented using any learning algorithm that creates a simple decision boundary between task-relevant images' scores or embeddings and those of non-target images, such as logistic regression or a support vector machine (SVM). In Sec. \ref{sec:results}, we show that cloud-feedback on thresholds outperforms non-adaptive schemes.}

\subsection{The lightweight \name sampling algorithm}
\label{subsec:lightweight_algorithm}

% OLD STUFF
%%%%%%%%%%%%%%%%%%%%%%%%%%%%%%%%%%%%%%%%%%%%%%%%%%%%%%%%%%%
\RSS{Having explained insights 1-3}, we
 can now present \name's sampler in Algorithm \ref{alg:harvestnet}, which serves as a \RSS{practical heuristic solution to Problem \ref{problem:rss_problem}}.
In each learning round $i$, a robot has a fixed perception model $\fDNNrobot^{i}$ and observes a stream of sensory inputs $\{x^{t,i}\}_{t=0}^{T}$. Sampling policy $\pisample$ stores sampled images in cache $\Cache^{i}$, which is uploaded to the cloud, annotated with ground-truth labels, appended to cloud dataset $\dataset^{i}$, and finally used to re-train perception model $\fDNNrobot^{i+1}$. This new model and an updated set of sampling parameters is then downloaded to the robot.
Learning repeats for a fixed set of rounds $\roundmax$, \RSS{until DNN validation accuracy saturates}.
Key steps of Algorithm \ref{alg:harvestnet} are colored.

\begin{algorithm}[h]
\DontPrintSemicolon
\SetNoFillComment % <---------------------------
\SetAlgoLined
 Assemble Targets $\targetclassset$, Cloud Dataset $\dataset^{0}$  \;

 Pre-train Initial Robot Perception DNN $\fDNNrobot^{0}$ on $\dataset^{0}$ \; 
 Init. $\{\yconfthresh^{0}, \rhothresh^{0}\} \leftarrow \updatethreshold(\dataset^{0}, \fDNNrobot^{0})$ \;

 \For{Learning Round $i~ \gets 0$ \KwTo $\roundmax$}{
    $\Cache^{i} \gets \{\emptyset\}$ \tcp*{Init Empty Cache}
    
    \For{$t~ \gets0$ \KwTo $T$}{
        $\boldy^t \gets [\yhat^t, \yconf^t, \emb^t] = \fDNNrobot^{i}(x^t)$ \;
        \HiLiGreen$a^t \gets \pisample(x^t, \boldy^t, \mathtt{TargSet}, \{\yconf_{\mathrm{thr.}}^{i}, \rho_{\mathrm{thr.}}^{i}\})$\; 
        $\Cache^{i}.\textsc{Append}(x^t) ~\mathbf{IFF}~ a^t = 1$\;
    }
    
    $\{(x^\tau, \yground^\tau)\} \gets \textsc{Annotate} (\Cache^{i})$ \; 
    
    $\dataset^{i+1} \gets \dataset^{i} ~\cup~ \{(x^\tau, \yground^\tau)\}, ~~\forall x^\tau \in \Cache^i$ \; 
    
    $\dtrain^{i+1}, \dval^{i+1} \gets \textsc{TrainValSplit} (\dataset^{i+1}) $ \; 
    
    \HiLiBlue$\fDNNrobot^{i+1} \leftarrow \gretrain(\fDNNrobot^i, \dtrain^{i+1})$ \tcp*{Re-train}\; 
    
    \HiLi$\{\yconfthresh^{i+1}, \rhothresh^{i+1}\} \leftarrow \updatethreshold(\dataset^{i+1}, \fDNNrobot^{i+1})$\;
    
 }
 \KwResult{Final DNN $\fDNNrobot^{\roundmax}$, Cloud Dataset $\dataset^{\roundmax}$}
 \caption{\name Sensory Stream Sampler}
 \label{alg:harvestnet}
\end{algorithm}

%\paragraph*{Comments on Algorithm}

\subsection{Benchmark Sampling Algorithms}
\label{subsec:benchmark_algorithms}
\noindent We compare \name with the following benchmarks: \\
\noindent \textbf{1. Random Sampling}: This scheme randomly fills the limited cache 
    from $T$ possible samples in round $i$.

\noindent \textbf{2. Non-adaptive cloud feedback}: 
This scheme does \textit{not} adapt thresholds using Eq. \ref{eq:update_threshold} (yellow line in Alg. \ref{alg:harvestnet}). 
\RSS{Rather, this scheme caches an image based on a probability that is proportional to the softmax confidence. For example, an image with a softmax confidence of 90\% is added to the cache with probability $0.90$. Analogously, an image with the 10\% quantile of embedding distances is cached with probability $0.90$, since lower embedding distances are preferred. Since this scheme does \textit{not} calibrate itself using a threshold, it often misses true positives which have low confidence scores at early rounds (Fig. \ref{fig:adapt_conf_threshold}, left)}.
We also experimented with an ``arg-max'' policy which stores an image if the confidence element for the target-class was most probable, but this performed poorly since the cutoff was too strict to store any task-relevant images.

\RSS{\noindent \textbf{3. \name-PriorityQueue}}: 
\RSS{This scheme caches an image into a priority queue ranked by either softmax confidence scores or embedding distances. The first items that are dropped when the priority queue is full are the target image with lowest softmax confidence or highest embedding distance.}

\noindent \textbf{4. Oracle Sampling}: 
This is an un-realizable upper bound since it \textit{perfectly} identifies images of the desired target at the robot. 
Since it is analytically hard to compute the \textit{best} set of limited images of size $\Cachesize$ to best improve the model, we instead approximate the upper bound by averaging several samples of $\Cachesize$ true target images, chosen \textit{without} any error.
\RSS{Importantly, the oracle takes the place of an ``all-cloud'' approach, since uploading all the data to the cloud is infeasible for network-limited robots like subterranean rovers.}

%%%%%%%%%%%%%%%%%%%%%%%%%%
\section{\name's experimental performance}
\label{sec:results}
\begin{figure}[t]
    \centering
    % revised for RSS 2020
    \includegraphics[width=1.0\columnwidth]{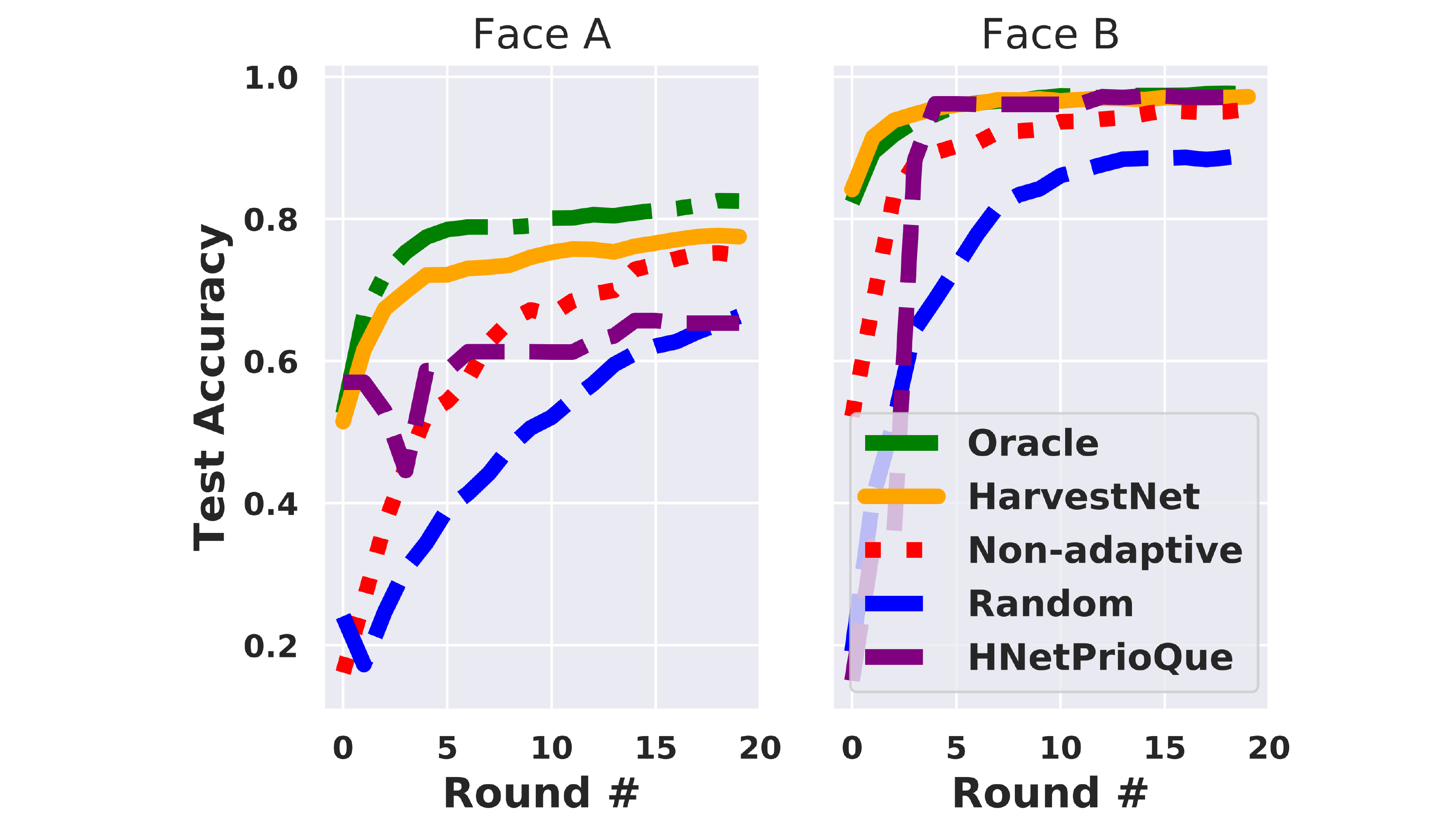}
    \caption{\textbf{FaceNet Results}: 
    We achieve steady accuracy gains
with minimal face examples, even matching the oracle performance for Face B, shown by the curve overlap.} 
    \label{fig:facenet_sampler_results}
\end{figure}

\begin{figure}[t]
    \centering
        \includegraphics[width=1.0\columnwidth]{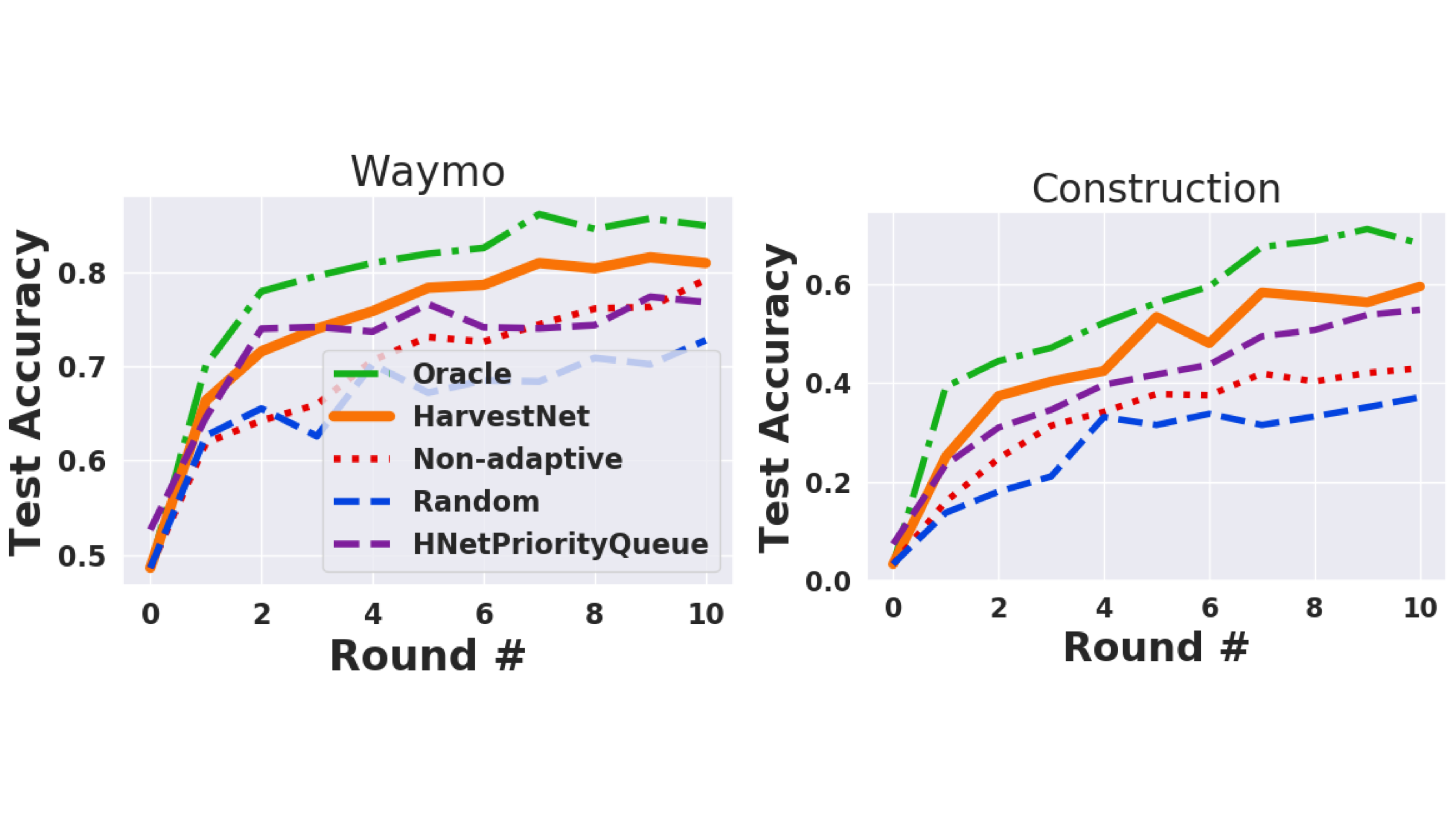}
    \caption{\textbf{Road-Scene Results}: \name outperforms 
    baselines that sample randomly or without cloud feedback.}
    \label{fig:road_scene_results}
\end{figure}

\noindent We now show \name's ability to (1)
achieve high model accuracy for a minimal number of sampled images (Figs. \ref{fig:facenet_sampler_results}, \ref{fig:road_scene_results}), and (2)
produce compute-efficient vision models. \RSS{Every benchmark that we evaluate has a fixed cache limit of $\Cachesize$ and thus obeys the constraints of Problem 1. Further, our results metric of high model accuracy is equivalent to minimizing model loss, which is the objective of Problem 1.}

% new figure
%%%%%%%%%%%%%%%%%%%%%%%%%%%%%%%%%%
\begin{figure}[t]
    \centering
    \includegraphics[width=1.0\columnwidth]{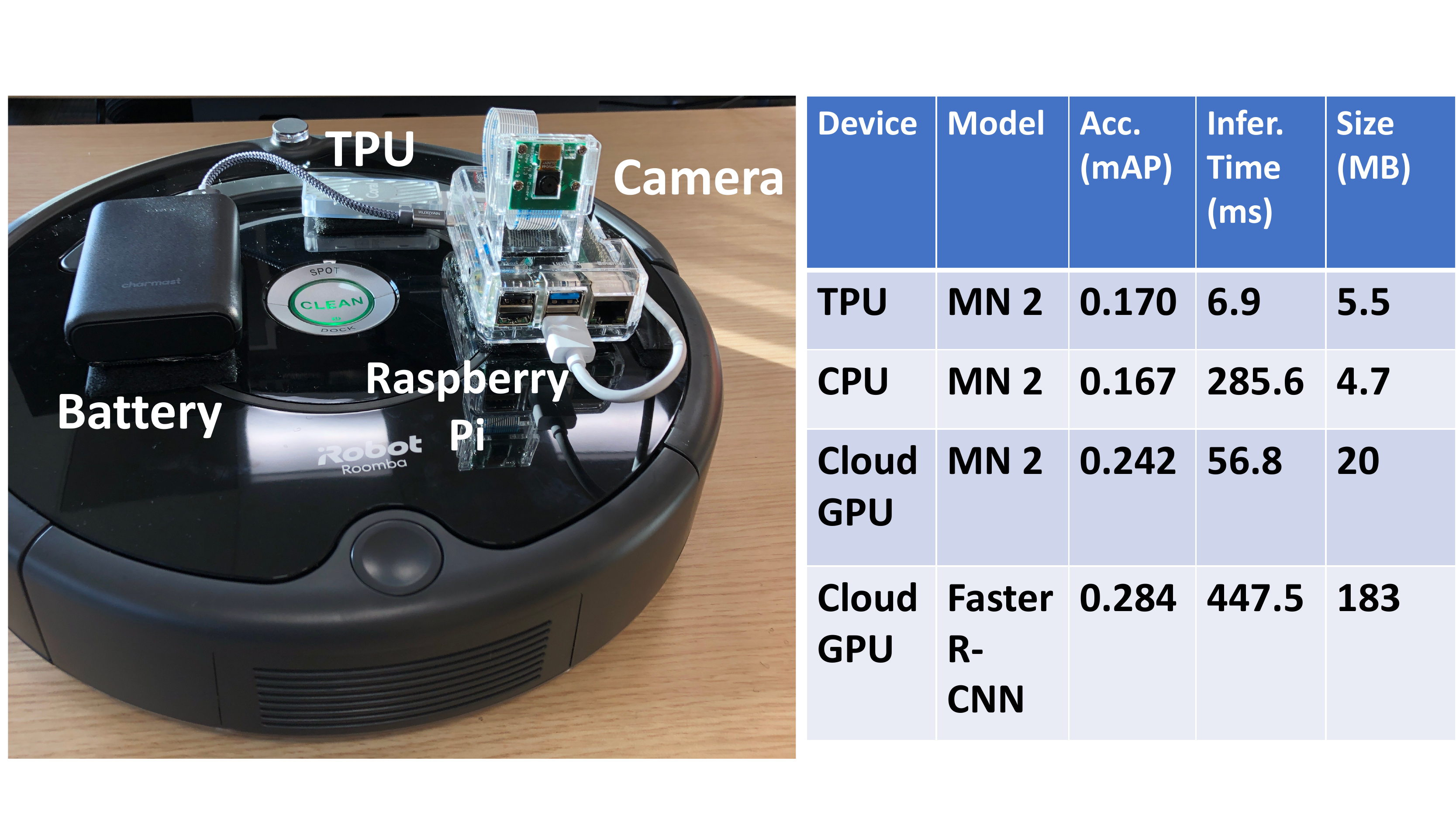}
    \caption{\textbf{Prototype}: (Left)
	\RSS{The lightweight sampler runs on an embedded CPU, such as on the Raspberry Pi,
	which uses an Edge TPU USB to accelerate DNN inference.}
    \textbf{Model Performance}: (Right) 
\name re-trained object-detectors flexibly run on the Edge TPU or cloud GPUs.}
    \label{fig:RPI_sampler}
\end{figure}

\subsection{Performance on Diverse Video Streams}

\paragraph*{FaceNet}
Our key result for the FaceNet scenario shows the final DNN classification accuracy on \textit{all test images} (y-axis) as more learning rounds $i$ progress (x-axis). 
As shown in Fig. \ref{fig:facenet_sampler_results} (left) our sampler of \name, in orange, achieves \SC{upto $1.68\times$} the final accuracy of benchmarks and no lower than \SC{$92\%$}
the accuracy of an upper-bound oracle solution in green. 
Fig. \ref{fig:facenet_sampler_results} also shows \name can flexibly adapt to different target faces of interest, since it shows steady model improvements for two randomly selected targets, referred to as Face A and B in the plots. Importantly, the robot is only provided $\targetset = 18$ initial examples of the face of interest and only caches a minimal set of images $\Cachesize = 10$ per learning round $i$, much less than the thousands of frames it observes per video.
We used the widely-adopted OpenFace \SC{\texttt{nn4.small2.v1}} DNN for face embeddings \cite{openface}, 
which were passed to an SVM for face-classification.

\paragraph*{Generalization to Dynamic Road Scenes}
We now stress test \name's ability to identify challenging road scenes of self-driving cars and construction sites. Both domains showcase \name's flexibility in using softmax confidence scores to decide if an image is likely a task-relevant training example to cache, unlike embedding distances in the FaceNet scenario. 
In each round $i$, we fine-tuned the Inception V3 DNN \cite{inception} in PyTorch on an Nvidia Tesla K80 cloud GPU for 20 epochs, which led to stable convergence, using hyper-parameters outlined in our technical report. 

Fig. \ref{fig:road_scene_results} shows the steady increase of final test set accuracy of various samplers (y-axis) for learning rounds $i$ in both the construction and self-driving car domains. 
\SC{For Waymo, \name, in orange, achieves upto \SC{$1.11\times$} the final accuracy of benchmarks and \SC{$93.5 \%$} the accuracy of an upper-bound oracle solution in green.} 
The most challenging test of \name comes from classifying whether a road scene has a construction site, since 
these rapidly changed daily. Still, \name outperforms non-adaptive sampling by over \SC{$1.38 \times$} times and is within \SC{87\%} of the oracle solution. 

\RSS{\paragraph*{Comparison with Priority Queue Version}
\name, which uses adaptive thresholds, convincingly outperforms benchmarks of non-adaptive
and random sampling. 
It also outperforms a variant of \name that uses the exact
same algorithm, but rather caches images in a Priority Queue, dubbed $\mathrm{HNetPriorityQueue}$, albeit by a lower margin such as in the Waymo domain. The priority queue version does worse when the proportion of target images per day is highly non-uniform, especially on days where there are no targets, leading it to cache irrelevant images below the empirical threshold.
Depending on how uniform the distribution of sensory inputs are per day, a designer might consider using the $\mathrm{HNetPriorityQueue}$ variant, which is also extremely compute-efficient and simple.}

\paragraph*{Benefits of re-training with field data}
To appreciate the accuracy gains of \name, we consider the final model accuracy if we trained a CNN \textit{solely} on high-quality public images of Waymo cars and construction sites and tested on our realistic road dataset. We could only find less than 200 Google Images for each scenario, which yielded an extremely poor accuracy of only \SC{41.6 \%} for Waymo and \SC{0 \%} for construction. Indeed, public images were often zoomed in, clear pictures from marketing images, and were not representative of real road scenes with partially occluded targets. 
Thus, our steady accuracy gains are substantial, especially with minimally-sampled images.
Unlike the rapid saturation of FaceNet and Waymo accuracy, the construction example
kept improving with time, serving as key benefit of continual learning. 

\paragraph*{Systems Bottleneck Savings}
The principal benefits of intelligent sampling derive from limiting the number of images cached in each learning round (e.g $\Cachesize = 60$). Thus, by re-training on between only \SC{18-34\%} of possible images, we decrease storage, annotation cost and time, and cloud-computing time by \SC{65.7-81.3\%}, since costs scaled linearly with the number of sampled images in our setup. 
    These estimates will vary considerably based on systems infrastructure.

\subsection{Prototype on Edge Tensor Processing Unit (TPU)}
\label{subsec:TPU_performance}

Finally, we show that training examples collected by \name can also be used for re-training object detection CNNs, shown in Fig. \ref{fig:contrast_retrain}, that scalably run on compute-limited robots.
Accordingly, we ran \name's DNNs on Google's Edge TPU development board \cite{edgeTPUwebsite},
where the TPU is a dedicated circuit for fast, low-power TensorFlow \cite{abadi2016tensorflow} model inference. The relatively new Edge TPU board supports, at best, a
quantized MobileNet 2 (MN2) DNN and features a complementary ARM Cortex embedded CPU, which runs DNNs without the aid of an AI accelerator.

Fig. \ref{fig:RPI_sampler} (right) shows that \name can flexibly re-train mobile-optimized models, such as MobileNet 2, or more accurate, but slower ``cloud-grade'' models such as \maskrcnn (contrasted in Fig. \ref{fig:contrast_retrain}). Thus, one can trade-off speed, size, and accuracy, quantified by the standard COCO mAP metric \cite{coco_detection_metrics}, for a wide variety of robot compute resources.

\RSS{For example, one can scalably run real-time vision algorithms on the TPU, which readily provides embeddings and class confidences, and pass them to \name's sampler running on an embedded CPU with minimal overhead. This configuration is shown on the robot we used in Fig. \ref{fig:RPI_sampler}, where the Raspberry Pi's CPU can decide whether to cache a new image with a mean speed of just $\approx .31$ms, which is an order of magnitude faster than DNN inference on the Edge TPU!}

\paragraph*{Limitations of our sampler}
\RSS{\name might fail in scenarios with extremely rare visual data, such as cases where we cannot obtain enough samples to seed the initial cloud dataset and train the initial robot model on round $0$. 
In future work, we can consider whether \textit{unknown weakpoints} of a model can be cached by storing examples with high confidence score entropy.}

%%%%%%%%%%%%%%%%%%%%%%%%%%
\section{Discussion and Conclusions}
\label{sec:discuss}

\RSS{The key contribution of this paper is to introduce a new, widely-applicable sensory sampling problem
to the robotics community and experimentally show the gains of our 
compute-efficient sampler. 
Since we anticipate the volume of rich robotic sensory data to grow, thereby making systems bottlenecks more acute, we encourage the robotics community
to significantly build upon our work. 
As such, we have open-sourced our dataset, Edge TPU vision DNNs, and sampling algorithm code at \url{https://sites.google.com/view/harvestnet}}.

In the future, we hope to enhance \name by 
fusing data from several cameras and LIDAR sensors on-board a robot. 
Further, the sampling algorithm can be modified to select adversarial training examples for vision DNNs.
\SC{Lastly, \name's sampler can be enhanced to explicitly not store ``private'' training data by rejecting such samples.}
%\SC{Lastly, we believe the \name architecture can be made to explicitly reject ``private'' training data by training modules (similar to a sampler) that do not store sensitive data, for example, on private property.}
We believe \name is a vital first step in keeping pace with surging sensory data in robotics by insightfully filtering video streams for actionable training examples, using insights from embedded AI, networking, and cloud robotics.

%% Use plainnat to work nicely with natbib. 
\balance
\bibliographystyle{plainnat}
\begin{small}
\bibliography{ref/ms.bib}
\end{small}

%\section*{Acknowledgments}
\newpage
\appendix
\section{Explanation of Systems Costs Calculations}

We now quantify system costs for cloud model re-training
for a small fleet of 10 robots in Table \ref{fig:system_cost_table}, each of which is assumed to capture 4 hours of driving data per day. In the first scenario, called ``Dashcam-ours'', we assume a single car has only one dashcam capturing
H.264 compressed video, as in our collected dataset. In the second scenario, called ``Multi-Sensor'', we base our calculations
off an Intel report \cite{intel} that estimates each self-driving car will generate 4 TB of video and LIDAR
data per day (1.5 hours of driving), of which video data alone is about 432 GB. 
%per car and day.

To calculate monthly storage cost, we use the Google Cloud Storage rate of \$0.026 per GB of monthly data \cite{GCS_pricing}. For network transfer time, we assume cars upload data over state-of-the-art, extremely fast 10Gb/s ethernet, when being serviced nightly at a garage.
To be conservative, we assume only 1\% of image data is interesting and needs to be annotated, specifically by the state-of-the-art Google Data Labeling service \cite{cloud_data_label}. 
We consider two cases that are: (1) these 1\% of video frames need to be annotated with bounding boxes at a standard rate of \$49 per 1000 boxes \cite{cloud_data_label_price}, or (2) these 1\% of frames need to be labeled with segmentation masks for \$850 per 1000 masks \cite{cloud_data_label_price}. On average, we assume each image has about 5 boxes or masks to be labeled. 
\SC{To roughly estimate annotation time, we hand-labeled 1000 bounding boxes in 1.4 hours. 
Using this rate, we estimate the number of days per month a team of 5 labellers will take in Table \ref{fig:system_cost_table}. Since we did not train models for segmentation, relevant estimates are marked as ``NA''}.

Overall, our estimates are conservative, since we assume a small fleet of 10 cars, extremely fast network transfer speeds, and pay for annotation of \textit{only 1\% of video data} because LIDAR annotation does not have
easily quantifiable costs from current data labelers.
%    and completely ignore the cost of LIDAR point cloud annotation, which currently does not have quantifiable costs from data labeling services. 
    %Nevertheless, Table \ref{fig:system_cost_table} shows the costs are quite significant, especially for human dataset annotation time and cost. 
    Nevertheless, Table \ref{fig:system_cost_table} shows significant costs, especially for dataset annotation. 
    %time and cost. 
%\SC{We emphasize our estimates are representative, but will vary widely depending on a company's available system resources}.
%\SC{which will vary depending on available resources.}

%% NON MULTI-COLUMN VERSION
\begin{table}[t]
    \centering
	\begin{tabular}{|l|l|l|l|l|}
		\hline
		\multirow{2}{*}{ \textbf{Bottleneck}} &
       \multicolumn{2}{c}{\textbf{Multi-Sensor}} &
       \multicolumn{2}{c|}{\textbf{Dashcam-ours}} \\
	     &  \textbf{B. Box} & \textbf{Mask} & \textbf{B. Box} & \textbf{Mask}\\ 
		\hline
		\cellcolor{Gray!20} Storage & \cellcolor{CornflowerBlue!20} \$31,200 & \cellcolor{YellowGreen!20} \$31,200  & \cellcolor{CornflowerBlue!20} \$172.50 & \cellcolor{YellowGreen!20} \$172.50 \\
		\hline
        \cellcolor{Gray!20} Transfer Time (hr) & \cellcolor{CornflowerBlue!20} 8.88  & \cellcolor{YellowGreen!20} 8.88 & \cellcolor{CornflowerBlue!20} 0.05 & \cellcolor{YellowGreen!20} 0.05 \\
		\hline
		\cellcolor{Gray!20} Annotation Cost & \cellcolor{CornflowerBlue!20} \$95,256 & \cellcolor{YellowGreen!20} \$1,652,400 & \cellcolor{CornflowerBlue!20} \$15,876 & \cellcolor{YellowGreen!20} \$275,400 \\
		\hline
        \cellcolor{Gray!20} Annot. Time (days) & 22 \cellcolor{CornflowerBlue!20} & NA \cellcolor{YellowGreen!20} & 18 \cellcolor{CornflowerBlue!20} & NA \cellcolor{YellowGreen!20} \\
		%\hline
		%\cellcolor{Gray!20} Cloud Computing & \cellcolor{CornflowerBlue!20} & NA \cellcolor{YellowGreen!20} & \cellcolor{CornflowerBlue!20} & NA \cellcolor{YellowGreen!20} \\
		\hline
	\end{tabular}
	%}
	\caption{\textbf{Systems Costs:}
    Monthly storage/annotation cost and daily network upload time are significant for self-driving car data even if we annotate \textit{1\% of video frames}.}
    %, as estimated above.}
	\label{fig:system_cost_table}
\end{table}

\end{document}